\documentclass[letterpaper, 10 pt, journal, twoside]{IEEEtran}

\IEEEoverridecommandlockouts                              

\usepackage{graphicx} 
\usepackage{times} 
\usepackage{amsmath} 
\usepackage{booktabs}
\usepackage{multirow}
\usepackage{siunitx}
\usepackage{url}
\usepackage{color}
\usepackage{float}

\newcommand{\rev}[1]{{#1}}
\newcommand{\omitifnotinarxiv}[1]{{#1}}
\newcommand{\statenet}{RAM Net}

\begin{document}

\title{Combining Events and Frames using Recurrent Asynchronous Multimodal Networks for Monocular Depth Prediction 
}
\author{\rev{Daniel Gehrig*,  Michelle R\"uegg*}, Mathias Gehrig,  Javier Hidalgo-Carri\'o, Davide Scaramuzza
\vspace{-2ex}
\thanks{Manuscript received: October 15th, 2020; Revised December 8th, 2020; Accepted February 13th, 2021.}
\thanks{This paper was recommended for publication by Editor Cesar Cadena Lerma upon evaluation of the Associate Editor and Reviewers' comments.}
\thanks{This work was supported by Prophesee, the National Centre of Competence in Research (NCCR) Robotics through the Swiss National Science Foundation (SNSF), and the
SNSF-ERC starting grant.}
\thanks{*These authors contributed equally. The authors are with the Robotics and Perception Group, University of Zurich, Switzerland (\protect\url{http://rpg.ifi.uzh.ch})}
}
\markboth{IEEE Robotics and Automation Letters. Preprint Version. Accepted February, 2021}
{Gehrig \MakeLowercase{\textit{et al.}}: Combining Events and Frames using Recurrent Asynchronous Multimodal Networks}
\maketitle
\begin{abstract}
Event cameras are novel vision sensors that report per-pixel brightness changes as a stream of asynchronous ``events".
They offer significant advantages compared to standard cameras due to their high temporal resolution, high dynamic range and lack of motion blur.
However, events only measure the varying component of the visual signal, which limits their ability to encode scene context. 
By contrast, standard cameras measure absolute intensity frames, which capture a much richer representation of the scene. 
Both sensors are thus complementary. 
However, due to the asynchronous nature of events, combining them with synchronous images remains challenging, especially for learning-based methods. 
This is because traditional recurrent neural networks (RNNs) are not designed for asynchronous and irregular data from additional sensors.
To address this challenge, we introduce Recurrent Asynchronous Multimodal (RAM) networks, which generalize traditional RNNs to handle asynchronous and irregular data from multiple sensors.
Inspired by traditional RNNs, RAM networks maintain a hidden state that is updated asynchronously and can be queried at any time to generate a prediction.
We apply this novel architecture to monocular depth estimation with events and frames where we show an improvement over state-of-the-art methods by up to 30\% in terms of mean absolute depth error. 
To enable further research on multimodal learning with events, we release EventScape, a new dataset with events, intensity frames, semantic labels, and depth maps recorded in the CARLA simulator.
\end{abstract}
\IEEEpeerreviewmaketitle

\section*{Multimedia Material}
Visit the project home page for code, dataset and more at
\url{http://rpg.ifi.uzh.ch/RAMNet.html}.
\section{Introduction}
\IEEEPARstart{E}{vent} cameras are bio-inspired vision sensors, that work radically different from conventional cameras. Instead of capturing the scene at a fixed frame rate, they report per-pixel brightness changes as a stream of asynchronous \emph{events} and do this with low latency and with microsecond resolution. 
The resulting stream of events, encodes the timestamp, pixel location and polarity (sign) of the brightness changes. 
\begin{figure}[t!]
  \centering
    \includegraphics[width=0.48\textwidth]{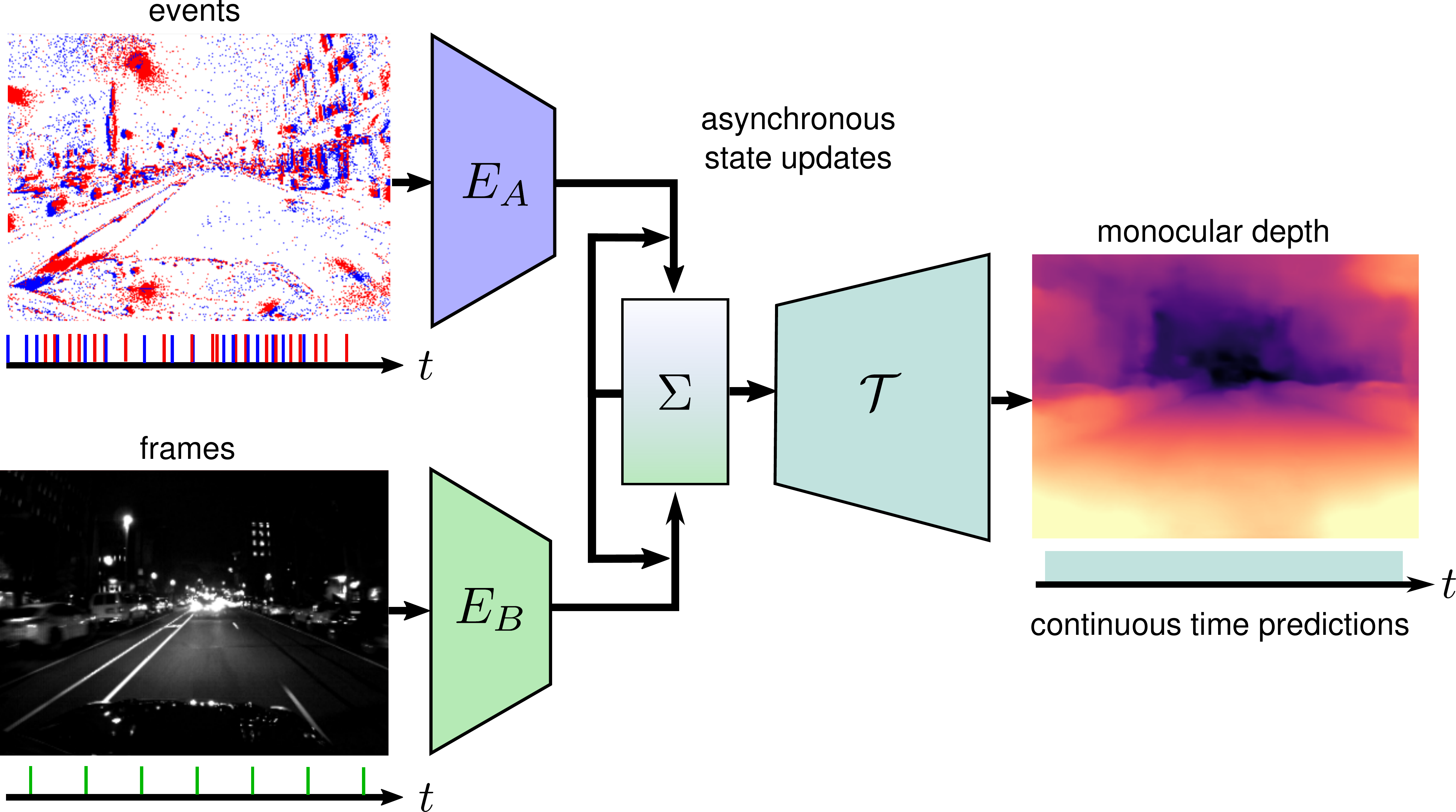}
  \caption{In this work, we introduce Recurrent Asynchronous Multimodal (RAM) networks for monocular depth estimation. Our method generalizes traditional recurrent neural networks (RNN) for asynchronous and irregular data from multiple sensors. It does this by maintaining an internal state $\Sigma$ that is updated asynchronously (through $E_A$ or $E_B$) and can be decoded to a task prediction $\mathcal{T}$ at any time instance. We apply this novel architecture to the task of monocular depth estimation from events and frames.}
  \label{fig:basic_architecture}
  \vspace{-3ex}
\end{figure}

Event cameras excel at sensing motion and do this at high speed and with low latency. 
However, compared to standard cameras that measure absolute brightness, they measure \emph{changes in brightness}. 
Event cameras are, thus, complementary to standard cameras, which motivates the development of novel computer vision algorithms which can combine the advantages of both sensors. 
For this reason, the Dynamic and Active-pixel Vision Sensor (DAVIS)~\cite{Brandli14ssc} was introduced. It combines a standard image sensor with an asynchronous event sensor on the same pixel array, making the sensor modalities synchronized and aligned.
More details about algorithms and applications of event cameras can be found in ~\cite{Gallego20pami}.

Despite these hardware alignments, designing algorithms for fusing both sensor modalities remains a challenging endeavor.
An ideal algorithm must take into account that events and frames are measured asynchronously and at different rates. 

While traditionally, filtering techniques, such as the Kalman Filter, have addressed this type of problem, they rely on accurate models designed by experts.
By contrast, data-driven sequence models such as LSTM ~\cite{Hochreiter97nc} directly learn a model from data and are thus capable of scaling to more complex problems. 
They are responsible for many of the recent breakthroughs in complex time sequence modeling but require that inputs be presented at fixed rates and synchronized with auxiliary inputs. 
For these reasons classical RNN-based methods are bound to fail when the input data violates these underlying assumptions ~\cite{Neil16nips,Chen18neurips}, as is the case for events and frames.

Methods for combining events and frames have taken two approaches. 
Model-based methods fuse events and frames by using a generative model that explains how events are triggered by the underlying brightness signal. 
However, due to nonlinearities and uncertainties in the event generation model, model-based fusion approaches 
are brittle and are sensitive to hyper-parameter tuning. 
For this reason, they have yet to demonstrate state-of-the-art performance on complex pixel-level tasks such as depth prediction~\cite{Zhu19cvpr} or image reconstruction~\cite{Rebecq19pami,Scheerlinck20wacv}. 
By contrast, learning-based methods have leveraged large datasets to generate more accurate predictions. 
However, current learning-based methods for events are limited in that they group events and frames into synchronized stacks which are passed to feed-forward neural networks ~\cite{Pini19iciap, Hu20arxiv}. 
Not only does this strategy sacrifice the asynchronicity and high temporal resolution of events, but it also limits the temporal context by using simple feed-forward networks instead of RNNs.
\rev{Instead, RNNs maintain a memory of previous states which encode temporal context over longer time windows and adapts it based on new inputs. Thus RNNs combine information from different views to guide depth prediction.} 

In this work we introduce Recurrent Asynchronous Multimodal (RAM) networks that leverage the complementarity of events and frames by \emph{(i)} explicitly taking into account the asynchronous and irregular nature of events and frames and \emph{(ii)} integrating
temporal contexts from several sensors through the use of recurrency. 
Inspired by traditional RNNs, RAM networks maintain an internal state that is updated asynchronously by either input modality and decoded to a prediction at any time (Fig. \ref{fig:basic_architecture}).  
We apply this novel network architecture to the task of event and frame-based monocular depth prediction which is an important building block in modern algorithms such as obstacle avoidance, path planning, and 3D mapping. 
By using event cameras these algorithms achieve increased robustness thanks to the advantages conferred by the sensor in high-speed and high-dynamic-range scenarios. 

Despite this interest, high-quality event and frame-based datasets for learning depth are scarce. 
To fill this gap we release we EventScape which we use to benchmark several baseline algorithms as well as RAM Net. 
EventScape is a new synthetic dataset that will enable further research on multimodal learning using events.
It is recorded in the CARLA~\cite{Dosovitskiy17corl} simulator and comprises events, intensity frames, semantic labels, depth maps and vehicle navigation parameters in diverse automotive scenes. 
Finally, we test our method on the multi-vehicle stereo event camera (MVSEC) dataset recorded with a real stereo DAVIS, where we show an improvement over state-of-the-art event-based methods of up to 30\% in terms of mean absolute depth error.
This is because depth predictions from RAM Net are more robust to changes in the frequency of input data.
In this work, we make the following contributions:
\begin{itemize}
\item We propose a new framework that can process multimodal inputs asynchronously and generalizes better to unseen asynchronous data compared to classical RNNs.
\item We apply our approach to the task of monocular depth estimation from events and frames, where we outperform state-of-the-art event-based methods by up to 30\%.
\item We release EventScape, a new multimodal dataset recorded in the CARLA simulator.  
\end{itemize}

\section{Related Work}
Due to the complementarity of events and frames, various algorithms have sought to leverage the advantages of both modalities by fusing them.
This trend has accelerated since the introduction of DAVIS in 2014~\cite{Brandli14ssc}.
Such algorithms include Simultaneous Localization and Mapping (SLAM)~\cite{Rosinol18ral}, feature tracking ~\cite{Kueng16iros,Gehrig18eccv,Gehrig19ijcv}, high-dynamic-range intensity reconstruction~\cite{Scheerlinck18accv} and image deblurring~\cite{Pan19cvpr}.
Although promising, these approaches depend strongly on the fidelity of the underlying models for combining events and frames.
Many algorithms today rely heavily on ideal sensor models~\cite{Gallego17pami, Gallego20pami}, which do not model noise or other dynamic effects and thus lead to performance degradations in non-ideal conditions. 
This is why for more complex tasks such as image reconstruction or monocular depth, state-of-the-art methods use a data-driven approach~\cite{Rebecq19pami,Zhu19cvpr,Tulyakov19iccv,Gehrig19iccv, Hidalgo20threedv}.
Of these, many rely on recurrent architectures which can leverage long time windows of events for improved prediction~\cite{Hidalgo20threedv,Rebecq19pami}.  
Although there exist many purely event-based learning methods, few address the fusion of images and events~\cite{Pini19iciap, Hu20arxiv,Alonso19cvprw}. These approaches fuse both modalities by synchronizing and concatenating both inputs and passing them to a standard feed-forward network ~\cite{Pini19iciap, Hu20arxiv,Alonso19cvprw}. 
While this strategy improves over passing each input individually, it discards the asynchronous nature and high temporal resolution of the events through stacking and synchronization. 
Moreover, fusion methods do not involve recurrency, further limiting the leveraged temporal context.

It is well known in the field of multimodal learning\footnote{In this paper we use multimodality to mean multiple data inputs, and do not refer to multimodal distributions used in statistics.} that traditional methods like RNNs struggle when sampling rates of different inputs is highly variable, and asynchronous~\cite{Chen18neurips,Neil16nips,Baltrusaitis2019ieee}.
Typical solutions involve padding, copying, or sampling rate conversions of the input, but at the cost of frequency reductions or temporal misalignments of input data~\cite{Oviatt2018}.  Alternative approaches use continuous-time models such as neural ODE's ~\cite{Chen18neurips} or phased LSTMs ~\cite{Neil16nips} to fuse such data but were so far only demonstrated on simple tasks.

RAM networks address all of these limitations by \emph{(i)} leveraging mature data-driven models such as RNNs for robust predictions on complex tasks such as monocular depth estimation, \emph{(ii)} maximizing the temporal context from events available  through the use of recurrency and \emph{(iii)} introducing individual and asynchronous state update rules for each input modality which retain the asynchronous nature and high temporal resolution of events.
\subsection{Monocular Depth Estimation with Events}
Monocular depth estimation is the task of estimating pixel-wise scene depth from a monocular input. 
Its applications range across different disciplines like medical imaging ~\cite{Chen17corr, Liu20tmi} or autonomous driving applications ~\cite{Laidlow19icra,Palafox19}. 
Several learning-based approaches based on frames have been proposed with promising results~\cite{Eigen14corr, Laidlow19icra, Godard18corr, Li18cvpr, Fu18corr, Wang17corr}. 
In particular, the use of event cameras for monocular depth estimation has become increasingly popular due to their unique properties, especially for autonomous driving, where low-latency obstacle avoidance and high-speed path planning is of particular importance. 
There exist different model-based approaches for purely event-based monocular depth estimation which predict sparse ~\cite{Rebecq17bmvc, Zhu17cvpr} or semi-dense depth maps ~\cite{Kim16eccv, Rebecq17ral}. 
Recently, various new learning-based methods have been introduced relying on feed-forward ~\cite{Tulyakov19iccv, Zhu19cvpr} and recurrent
architectures~\cite{Hidalgo20threedv}. 
Of these methods, all are purely event-based and thus cannot take advantage of complementary frames which can provide important queues in scenarios with little to no ego-motion, where few events are triggered.
By contrast, our method, combines frames and events thus leveraging the advantages of both.  
It does this by generalizing the recurrent architecture in ~\cite{Hidalgo20threedv,Rebecq19pami} to handle asynchronous and irregular inputs from both sensors.

\section{Method}

\subsection{Recurrent Asynchronous Multimodal Networks}
The working principle of RAM Networks is shown in Fig. \ref{fig:network_architectures}, where we consider the task of monocular depth prediction from events and frames. 
Let us first consider a scenario where we want to fuse the data from $N$ sensors which provide measurements at steadily increasing time stamps $t_j$.
We may assign a sensor $k_j\in\{1,2,...,N\}$ and measurement $x_{k_j}(t_j)$ from this sensor to each time $t_j$. 
We are thus faced with fusing a stream of measurements $\{x_{k_j}(t_j)\}_{j=1}^T$.
\rev{Since the measurements from different sensors may have vastly different formats, RAM Networks first maps the stream of measurements to a stream of \emph{intermediate features}, $s_{k_j}$ through the use of sensor specific learnable encoders $\mathbf{E}_{k_j}$
\begin{equation}
    s_{k_j} = \mathbf{E}_{k_j}\left(x_{k_j},\theta_{k_j}\right).
\end{equation}
Where $\theta$ denote the network parameters of RAM Net.}
In this work we collect intermediate features at several scales of the encoder, as can be seen in Fig. \ref{fig:network_architectures}.
The resulting stream $\{s_{k_j}(t_j)\}_{j=1}^{T}$ has the following properties: (i) it is asynchronous, which means that features from different sensors may appear in any order in the sequence and (ii) the data rates are variable, which means that the time interval between features changes over time. 
We, therefore, need to make sure to fuse these features correctly. 
\rev{To retain both these properties we use sensor-specific state combination operators $h_{k_j}$ which take a feature $s_{k_j}$ from sensor $k_j$ and use it to update the latent variable $\Sigma_{j-1}$ at time $t_{j-1}$}:
\begin{equation}
    \Sigma_j = h_{k_j}(\Sigma_{j-1},s_{k_j},\theta).
\end{equation}
The combination of features with latent variable can be performed via simple summation, concatenation or with more elaborate neural network layers. 
In this work, we choose to combine the states with a convolution gated recurrent unit (convGRU)~\cite{Siam17icip} for each sensor and at each scale, which is depicted as a black square in Fig. \ref{fig:network_architectures}.
This is important for the use of encoder-decoder architectures with skip connections. 
The update equation thus reads:
\begin{align}
    \tilde\Sigma_j&=\Phi(r\odot \Sigma_{j-1},s_{k_j})\\
    \Sigma_j &= (1-z)\odot\Sigma_{j-1}+z\odot \tilde\Sigma_j\\
    \nonumber\text{with }&     z=f(\Sigma_{j-1},s_{k_j})\text{ and }
    r=g(\Sigma_{j-1},s_{k_j})
\end{align}
Where $f$ and $g$ denote single convolutions followed by a sigmoid and $\Phi$ denotes a convolution followed by $\text{tanh}$. 
By using dedicated state combination operators we ensure that features from different sensors are first mapped to a suitable space such that they can be easily fused with $\Sigma_j$.
The state combination operators are thus applied sequentially, alternating between different sensors, as their measurements become available.
Due to the recursive update of the state $\Sigma_j$, it always contains information of both modalities, independent of which modality was seen last. 
This produces a stream of latent variables $\{\Sigma_j\}_{j=1}^{T}$.
These can be decoded to the task variable $y_j$ at each timestep $t_j$ with the task decoder $\mathcal{T}$
\begin{equation}
    y_j = \mathcal{T}(\Sigma_j,\theta)    
\end{equation}
Since the $y_j$ only depends on $\Sigma_j$, RAM Networks are Markovian, i.e. they are independent of the history of sensor measurements, given $\Sigma_j$. 
Thus $\Sigma_j$ encodes the history of measurements.
There is a strong connection with hidden Markov models (HMMs), which similarly model a latent variable given sensor measurements and impose the same conditional independence as RAM Networks.
In the following section, we will explore how this framework is applied to the task of event and frame-based depth estimation.

\subsection{Network Architecture for monocular depth estimation}

\begin{figure}[tpb]
  \centering
    \includegraphics[width=0.43\textwidth]{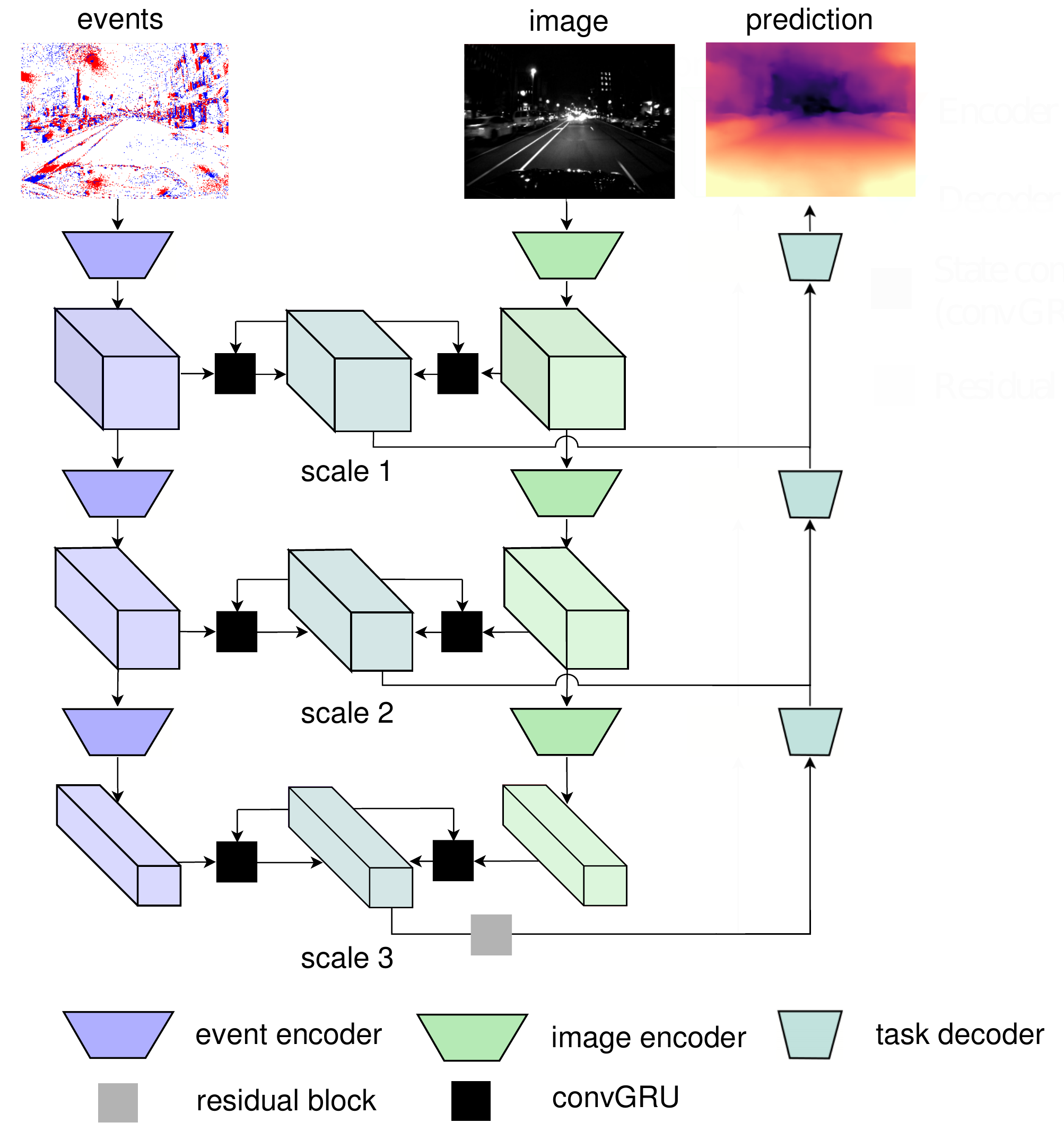}
  \caption{Network architecture of \statenet{} for monocular depth estimation with events and frames. The architecture is a fully convolutional encoder-decoder architecture based on U-Net ~\cite{Ronneberger15icmicci}. Inputs can be processed asynchronously. Predictions are dependent on all previous inputs due to the recursive calculation of the state.}
  \label{fig:network_architectures}
  \vspace{-2ex}
\end{figure}

Fig. \ref{fig:network_architectures} shows in detail how the general network architecture is adapted for this task. The basic architecture is inspired by U-Net ~\cite{Ronneberger15icmicci}, which was previously used for monocular depth estimation with events ~\cite{Hidalgo20threedv}. 
At each scale, we use the skip connection as intermediate features $s_{k_j}$. 
The intermediate features $s_{k_j}$ are combined with $\Sigma_{j-1}$ via a ConvGRU layer to produce $\Sigma_j$.
The latent variable at the lowest scale is directly fed into a residual block, followed by three decoder levels. 
At the skip connections, the decoder input is combined with the state of the corresponding stage by summation. 
Each encoder consists of a down-sampling convolution with kernel size 5 and stride 2. 
The residual block is composed of two convolutions with kernel size 3 and the decoders use bilinear upsampling followed by a convolution with kernel size 5. 
All layers use a ReLU  activation function. 
\omitifnotinarxiv{More architecture details can be found in Tab. I and Sec. 1-C of the appendix.}

\subsection{Event Generation Model}
Before we see the application of RAM Networks to events and frames, we review the working principle of the event camera.
Event cameras contain independent pixels $\mathbf{u}$ that react to changes in the log brightness signal $L(\mathbf{u},t)$. 
If the magnitude of the log brightness changes by more than a threshold $C$ since the last event, a new event $e_k = (x_k, y_k, t_k, p_k)$ is triggered at that pixel location $\mathbf{u} = (x_k, y_k)^T$. 
The polarity $p_k$ takes the values $\{-1, +1\}$, depending on the direction of brightness change. 
The generative event model for an ideal sensor ~\cite{Gallego17pami, Gallego20pami} states that an event with polarity $p_k$ is triggered at pixel $\mathbf{u}_k$ and timestamp $t_k$ when:
\begin{equation}
\label{eq:gen_model}
    \Delta L(\mathbf{u}_k, t_k) = p_k\left(L(\mathbf{u}_k, t_k) - L(\mathbf{u}_k, t_k - \Delta t_k)\right) \geq C.
\end{equation}
Here $\Delta t_k$ is the time since the last event at the same pixel.
\subsection{Event Representation}
To leverage existing convolutional neural networks architectures, events are typically converted to a fixed-size tensor. 
A set of input events $\epsilon = \{e_i\}_{i=0}^{N-1}$ in the time window $\Delta T = t_{N-1} - t_0$ is and drawn into a voxel grid with spatial dimensions $H\times W$ and $B$ temporal bins ~\cite{Rebecq19pami, Zhu19cvpr,Gehrig19iccv}. 

\begin{equation}
\mathbf{V}(x,y,t) = \sum_i p_i \delta(x-x_i,y-y_i) \max\{0,1-|t-t_i^*|\},
\end{equation}
Here $t_i^* = \frac{B-1}{\Delta T}(t_i-t_0)$.
As in ~\cite{Rebecq19pami}, the number of bins $B$ is set to 5 for all experiments.

\subsection{Depth Representation}
The metric depth $\widehat{D}_{m,k}$ is first converted into a normalized log depth map $\widehat{D}_k\in[0,1]$ which facilitates learning of large depth variations~\cite{Hidalgo20threedv, Li18cvpr, Eigen14corr2}. 
The maps are constructed pixel-wise. 
The logarithmic depth is calculated as
\begin{equation}
    \widehat{D}_k = \frac{1}{\alpha}\log\frac{\widehat{D}_{m,k}}{D_{\text{max}}}+1,
\end{equation}
where $\alpha$ is chosen such that the closest observed depth is mapped to 0. 
For EventScape we choose $\alpha=5.7$ and $D_{\text{max}}=\SI{1000}{\metre}$ whereas for MVSEC we choose $\alpha=3.7$ and $D_{\text{max}}=\SI{80}{\metre}$. 
For monocular depth estimation, the depth can only be estimated up to scale. 
Some methods circumvent this problem by using additional information from a stereo setup, IMU data, or camera poses ~\cite{Tulyakov19iccv, Rosinol18ral, Zhu17cvpr, Rebecq16bmvc}. 
However, using a recurrent deep learning architecture and manually setting the maximum depth observed in the dataset leads to satisfying results without additional data.

\begin{figure*}[t!]
    \centering
    \begin{tabular}{cccc}
        \includegraphics[width=0.22\linewidth]{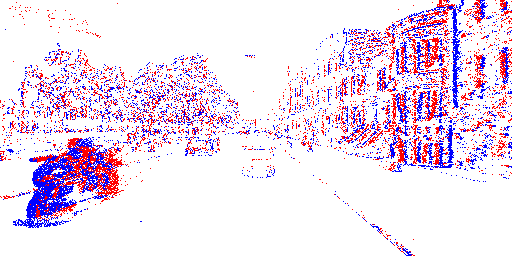}&
        \includegraphics[width=0.22\linewidth]{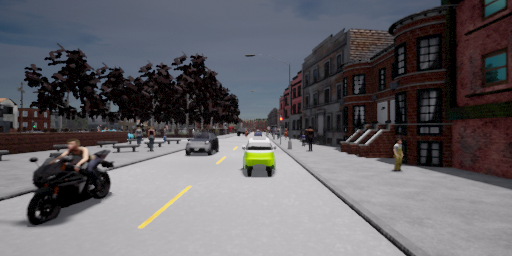}&
        \includegraphics[width=0.22\linewidth]{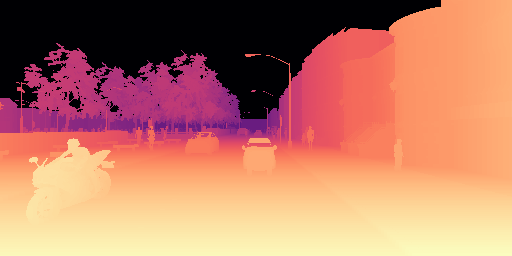}&
        \includegraphics[width=0.22\linewidth]{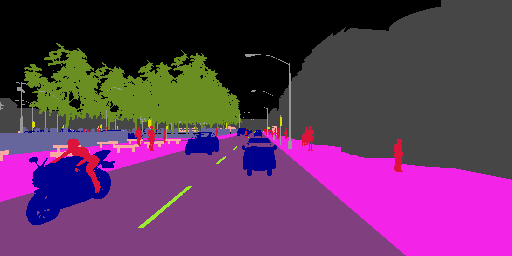}
        \\
        \footnotesize{(a) events} & \footnotesize{(b) images}& \footnotesize{(c) depth maps}& \footnotesize{(d) segmentation labels}
    \end{tabular}
    \caption{Sample data from the EventScape dataset. 
    It contains events (a), images (b), depth maps (c), segmentation labels (d), and vehicle navigation parameters. 
    Images, depth maps, and segmentation labels are provided at 25Hz while vehicle navigation parameters are provided at 1000Hz. 
    Events are generated by converting 500Hz images to events using the event camera plugin in CARLA~\cite{Hidalgo20threedv} which is based on the event camera simulator~\cite{Rebecq18corl}.
    We release this dataset to enable further research in multimodal learning with events.}
    \label{fig:eventscape}
    \vspace{-3ex}
\end{figure*}
\subsection{Training}
RAM Net is trained in a supervised fashion by using ground truth depth map labels provided at sparse time points. 
In the case of EventScape, these labels are provided by CARLA and for MVSEC these are measured using a LiDAR. 
We use a combination of a scale-invariant loss and a multi-scale scale-invariant gradient matching loss which we compute for valid ground truth labels and sum over a sequence of events and images.
The scale invariant loss ~\cite{Eigen14corr2} is defined as: 
\begin{equation}
    \mathcal{L}_{k,si} = \frac{1}{n} \sum_\mathbf{u} (R_k(\mathbf{u}))^2 - \frac{1}{n^2}\left(\sum_\mathbf{u} R_k(\mathbf{u})\right)^2,
\end{equation}
with $n$ counting the valid ground truth pixels.
The multi-scale scale-invariant gradient matching loss ~\cite{Li18cvpr} is defined as:
\begin{equation}
    \mathcal{L}_{k,grad} = \frac{1}{n} \sum_s\sum_\mathbf{u} |\nabla_x R_k^s(\mathbf{u})| + |\nabla_y R_k^s(\mathbf{u})|.
\end{equation}
Where $\nabla_x$ and $\nabla_y$ compute the edges in the x and y direction respectively by using the well-known Sobel operator.
The loss is calculated over four different scales, indicated with $s$. 
It favors smooth gradient changes and enforces sharp depth discontinuities in the prediction ~\cite{Li18cvpr} since it tries to match the gradients in the ground truth depth.
The resulting total loss for a sequence of length $L$ is
\begin{equation}
    \mathcal{L}_{tot} = \sum_{k=0}^{L-1} \mathcal{L}_{k,si} + \lambda\mathcal{L}_{k,grad}.
\end{equation}
The weight of the gradient loss $\lambda$ is chosen as $0.25$.
To facilitate learning, the input data is normalized, cropped randomly to $224 \times 224$ and we apply random horizontal flipping. 
Non-zero entries in the voxel grid normalized such that their mean and variance are 0 and 1 respectively.  

The network is implemented in PyTorch ~\cite{Paszke17nipsw} and optimized using the ADAM optimizer ~\cite{Kingma15iclr} with a batch size of 8. 
On EventScape, we choose a learning rate of 0.0003 and train for 27'000 iterations. 
For MVSEC on the other hand, we use a learning rate of 0.001 and train for 7'600 iterations.
\section{Experimental Setup}
In a first step, we validate RAM Net on the synthetic EventScape dataset where we compare against event-only, frame-only, and event and frame-based baselines.
We show that our method not only achieves better performance compared to unimodal baselines but also outperforms multi-modal baselines.
We then show results on the real world MVSEC dataset, where we compare against state-of-the-art frame-based~\cite{Li18cvpr} and event-based~\cite{Hidalgo20threedv,Zhu19cvpr} methods as well as our image-only baseline.
We show that RAM Net outperforms state-of-the-art methods by up to 30\% in terms of absolute relative depth error.
\subsection{Datasets}
\noindent\textbf{EventScape:}\label{sec:eventscape}
EventScape is a large scale, synthetic dataset recorded in the CARLA simulator ~\cite{Dosovitskiy17corl}. 
It contains 743 sequences of driving data with a total of 171'000 labels, equivalent to around 2 hours of driving. 
It is recorded in different cities with different dynamic actors like pedestrians and cars. 
In addition to events and image data, we recorded depth, semantic segmentation, and various vehicle control data.
An example from the dataset can be viewed in Fig. \ref{fig:eventscape}.
While frames, depth, and semantic labels are provided synchronously at 25Hz, vehicle control data is recorded at 1000 Hz.
Events are generated by using the event camera plugin in CARLA~\cite{Hidalgo20threedv} which renders images at 500Hz and converts them to asynchronous events using the event camera simulator ESIM ~\cite{Rebecq18corl}.
ESIM generates events by randomly sampling a contrast threshold from the range $[0.15,0.25]$ for each sequence and applying Eq.\eqref{eq:gen_model} to the 500Hz frames. 
To improve the realism of the events we include a refractory period $\SI{100}{\micro\second}$ which is similar to that observed in real sensors.
Furthermore, this procedure was employed in previous works~\cite{Rebecq19pami}, where it was shown to boost performance.
The events and frames are synchronized and aligned, resembling the output of a DAVIS sensor ~\cite{Brandli14ssc}. 
For training, CARLA towns 1, 2, and 3 are used, validation and test data are drawn from town 5 from geographically separated areas. 
\newline
\newline
\noindent\textbf{MVSEC:}\label{sec:mvsec}
The multi-vehicle stereo event camera dataset (MVSEC)~\cite{Zhu18ral} consists of data recorded using two DAVIS event cameras with a resolution of $346\times260$ pixels in a stereo rig, together with measurements from a LiDAR for ground truth depth.
It features several day and night-time driving sequences as well as indoor sequences recorded on a quadcopter. 
The depth maps were recorded at 20Hz while the grayscale images were recorded at 10 Hz for night time and 45 Hz for daytime sequences. 

\rev{\subsection{Baselines}
To understand the impact of multimodality and the asynchronous update scheme employed by RAM Net we implement} three recurrent baselines termed \emph{(i)} purely event-based (E), \emph{(ii)} purely frame-based (I) and \emph{(iii)} event and frame-based (E+I). \rev{Of these baselines E and I are designed to understand the impact of multimodality, while E+I ablates the proposed asynchronous update scheme.}
They have identical residuals and decoders but instead of recurrent state combination operators, they feature recurrent convLSTM encoders at each level.
\omitifnotinarxiv{For a more detailed overview of the baseline architectures, see Tab. II and Sec. 1-C in the appendix.} 
While E only receives voxel grids as input, I receives only gray-scale frames and E+I receives stacks of voxel grids and frames, similar to ~\cite{Pini19iciap}.
For E+I when a new voxel grid arrives, we stack it with a copy of the last seen image. 
Of the described baselines E+I has access to the same amount of information as our method, but there is a temporal mismatch between event and image data.
Supervision is performed sparsely, indicated with a blue arrow in Fig. \ref{fig:plots_statenet_vs_baselines}. 
\rev{Finally, to understand the effect of recurrency we compare against a simple event- and frame-based feed-forward network, based on \cite{Pini19iciap} which we term E+I no recurrency. 
Here we replace the convLSTM blocks with simple convolutions and train on 200 ms of events stacked with the corresponding images at 5 Hz.}
\section{Results}
\subsection{Effect of Input Modality and Architecture}
\label{sec:regular_asynchronous_experiments}
\begin{figure*}[thpb]
    \centering
    \begin{tabular}{ccc}
        \includegraphics[width=0.32\linewidth]{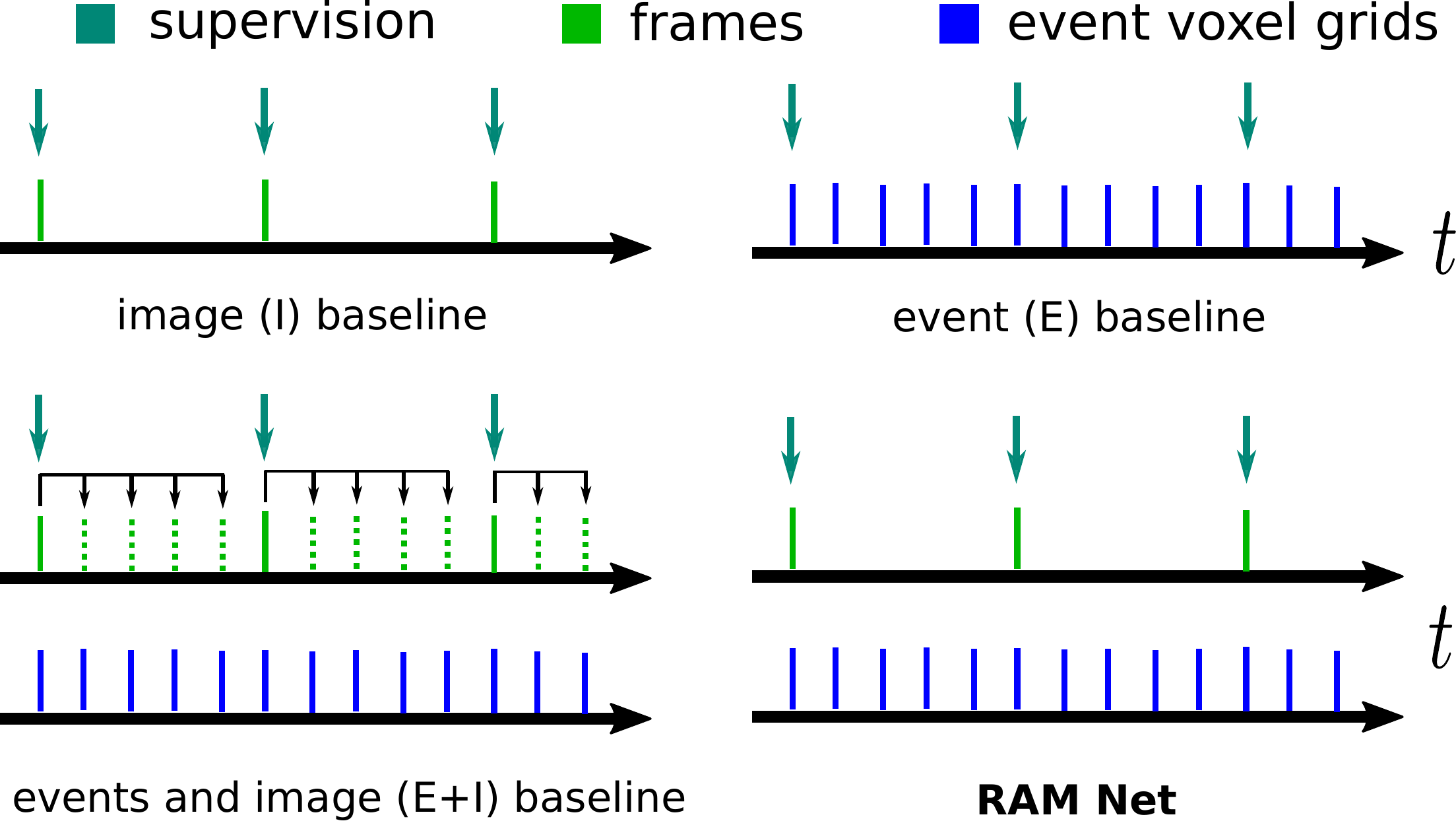}&
        \includegraphics[height=0.2\linewidth]{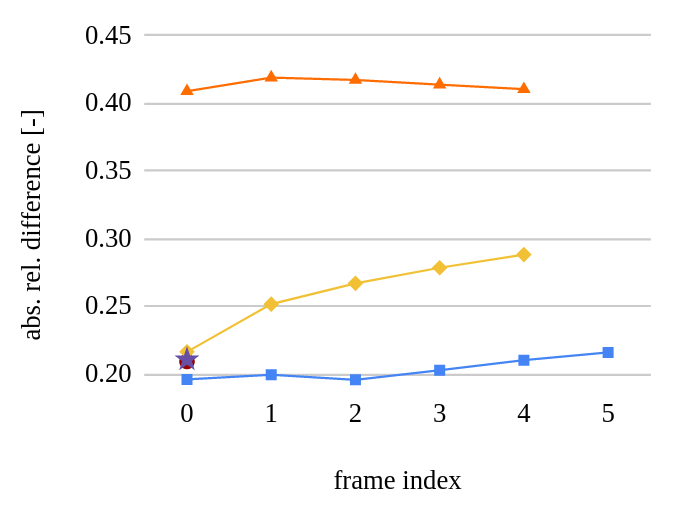}&
        \includegraphics[height=0.2\linewidth]{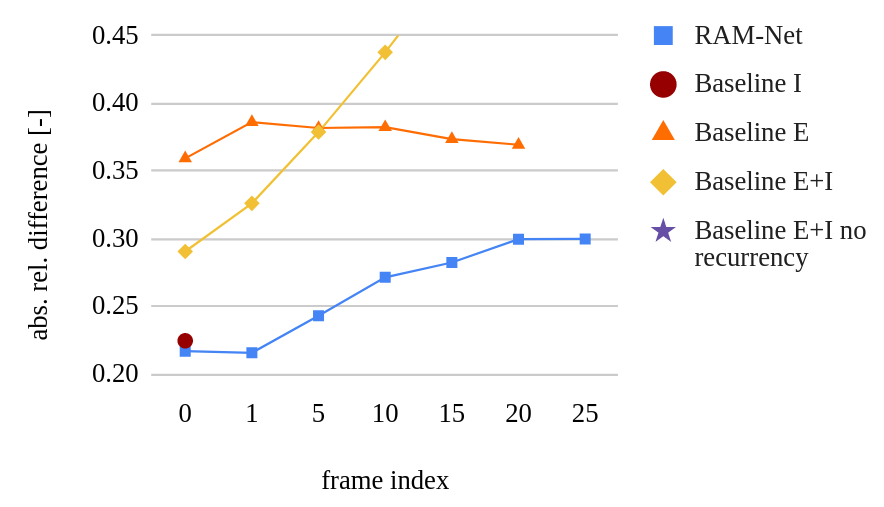}\\
        \footnotesize{(a) baselines and our method}&\footnotesize{(b) Images at 5Hz, events at 25Hz} & \footnotesize{(c) Images at 1Hz, events at 25Hz}
    \end{tabular}
    \caption{Comparison of \statenet{} and baselines (a). \statenet{} outperforms the baselines in all data points. The x-axis (b-c) depicts the input to the network before the prediction that was used to calculate the corresponding data point. (b) Images occur at 5 Hz, depth labels and event data is available at 25Hz, which we use to train network. (c) Generalization study with images at 1Hz and events at 25 Hz, as during training.}
    \label{fig:plots_statenet_vs_baselines}
    \vspace{-3ex}
\end{figure*}
As a first step, we train all baselines and RAM Net with 5 Hz images and depth labels, and voxel grids constructed at 25 Hz on EventScape. 
We construct sequences from 10 sub-sequences, each containing 5 voxel grids followed by one frame and one depth label.
We then pass these sub-sequences at each step of the sequential model unrolling, resulting in 10 frames and 50 voxel grids per sequence. 
A loss is then computed after each depth label, i.e. at 5Hz.
After each sub-sequence, RAM Net predictions after the last voxel grid and after the subsequent image are temporally aligned. 
We therefore average the per-pixel loss for both predictions, which we found improved training.
By contrast, the baselines only see a single loss.
\newline
\newline
\noindent\textbf{Results:} The results after computing the absolute relative depth error~\cite{Li18cvpr} can be seen in Fig. \ref{fig:plots_statenet_vs_baselines} (b). 
On the x-axis, a frame index 0 indicates that the metric was averaged over image predictions.
For E we evaluate the prediction from the most recent voxel grid.
The E+I baseline sees the concatenation of the new image and the current events.
For frame index $n$ the metric was computed on predictions after $n$ event tensors. 
As the I and E+I recurrent baselines are only able to predict at image frame rate, it is the only configuration that is represented as a single data point in Fig. \ref{fig:plots_statenet_vs_baselines}. 
Baselines E and E+I predict an output after each event voxel grid or combined event-image input. 
As \statenet{} outputs separate predictions for the events and image inputs, it contains an additional data point in Fig. \ref{fig:plots_statenet_vs_baselines}, indicated as frame index 5.

After the frame prediction, \statenet{} outperforms all baselines. 
As we move further from the initial frame, the prediction error gracefully increases but remains at lower levels than methods E and E+I (\rev{49\% and 38\% reduction respectively in terms of absolute relative difference at frame index 4 in Fig. \ref{fig:plots_statenet_vs_baselines} (b))}.
\rev{Compared to \statenet{}, the I baseline shows a 5\% higher difference, but cannot generate predictions at a high temporal resolution as the other methods.} 
Furthermore, while E+I has a strong prediction after a new image input the prediction error increases due to the gradual misalignment between events and copied frames.
The E baseline has the worst performance but has constant metrics over time since it is independent of the last seen image 
\rev{Finally, E+I no recurrency at 0.211 exhibits a slightly lower error at frame index 0, than its recurrent counterpart, E+I, at 0.216. This small difference likely comes from the fact that, for E+I, events and images become more temporally misaligned, as the frame index increases, leading to a higher error (Fig. \ref{fig:plots_statenet_vs_baselines} b).
Upon seeing a new image, the network learns to correct this error but fails to do so completely. 
By contrast, the E+I no recurrent baseline does not suffer from this misalignment and thus performs slightly better.
However, crucially, this comes at the cost of a lower prediction rate compared to the E+I recurrent baseline (5 Hz vs. 25 Hz).
RAM Net outperforms both methods with an error of 0.198 by leveraging the advantages of both. 
It does not suffer from temporal misalignment of inputs since images are never copied and is also able to make predictions between frames. 
These experiments show \emph{(i)} the benefit of combining frames and events which highlights their complementarity, \emph{(ii)} the effect of temporal alignment for recurrent methods and \emph{(iii)} the importance of recurrency which enables high-quality depth predictions in the blind-time between image frames.}

\subsection{Generalization to Different Data Rates}
To further study the network behavior, we analyze its ability to generalize to new frame rates. 
For the experiment in Fig. \ref{fig:plots_statenet_vs_baselines} (c) we use the networks trained on 5Hz image data and tested them on 1Hz image data.
Event data was still available at 25 Hz.
This task is much more challenging since the gaps between images become larger which means that networks need to be able to retain a longer memory.
\newline
\newline
\textbf{Results:} The I baseline performs slightly worse, which can be explained by the increased time interval between seen images. 
The small decrease in performance indicates that I does not heavily depend on recurrency. 
The prediction quality after the image input also decreases for \statenet{}. 
As before, the accuracy of \statenet{}'s predictions gracefully degrades for increasing frame index, reaching a plateau at index 20. 
Nonetheless, \statenet{} retains a low error compared to E+I and E. 
This time, the error for E+I grows drastically, even surpassing E. 
This is because for large time gaps the events and copies of the previous frame show significant misalignment.
This shows that \statenet{} retains additional information from the image even after a large time-span.
Furthermore, this shows that \statenet{} can generalize well to data inputs at different frequencies. 
\omitifnotinarxiv{For a study on the sensitivity of our network to different inputs, see Fig. 1 and Sec. 1-B in the appendix.}
\subsection{Real World Experiments}
Next we evaluate our method on MVSEC. 
Since ground truth depth and images are no longer synchronized, labels and network predictions need to be temporally aligned. 
To do this, we split the events between depth labels into equal packets of 10'000 events, distributing the surplus of events equally, if required.
Next, we generate voxel grids from packets, with the timestamp corresponding to that of the last event in the packet. 
We feed this irregular sequence of voxel grids to the network together with frame sequence, alternating between frames and voxel grids as they appear. 
\begin{table*}[t!]
\centering
\caption{Average absolute depth errors (in meters) at different maximum cut-off depths.} 
\label{tab:results_mvsec}
\resizebox{0.87\textwidth}{!}{%
\setlength{\arrayrulewidth}{.1em}
\begin{tabular}{c|c|ccc|cc|cc}
\hline
\multicolumn{1}{c|}{\multirow{2}{*}{\textbf{Dataset}}} & \multicolumn{1}{c|}{\multirow{2}{*}{\textbf{Cutoff}}} & \multicolumn{3}{c|}{\textbf{frame-based}}                                                                                                 & \multicolumn{2}{c|}{\textbf{event-based}}                                       & \multicolumn{2}{c}{\textbf{event and frame-based}}                                        \\
\multicolumn{1}{c|}{}                                  & \multicolumn{1}{c|}{}                                 & \multicolumn{1}{c}{MegaDepth ~\cite{Li18cvpr}} & \multicolumn{1}{c}{I baseline {[}S{]}} & \multicolumn{1}{c|}{I baseline {[}S$\to$  R{]}} & \multicolumn{1}{c}{Zhu et al. ~\cite{Zhu19cvpr}} & \multicolumn{1}{c|}{E2Depth ~\cite{Hidalgo20threedv}} & \multicolumn{1}{c}{RAM Net {[}S{]}} & \multicolumn{1}{c}{RAM Net {[}S$\to$ R{]}} \\ \hline
                                                       & \SI{10}{\meter}                                                   & 2.37                                   & 13.02                                  & 1.74                                                    & 2.72                                    & 1.85                                  & 3.77                                & \textbf{1.39}                                       \\
outdoor day1                                           & \SI{20}{\meter}                                                   & 4.06                                   & 11.37                                  & 2.55                                                    & 3.84                                    & 2.64                                  & 4.27                                & \textbf{2.17}                                       \\
                                                       & \SI{30}{\meter}                                                   & 5.38                                   & 11.29                                  & 3.07                                                    & 4.40                                    & 3.13                                  & 5.17                                & \textbf{2.76}                                       \\ \hline
                                                       & \SI{10}{\meter}                                                   & 2.54                                   & 7.93                                   & 2.72                                                    & 3.13                                    & 3.38                                  & 4.00                                & \textbf{2.50}                                       \\
outdoor night1                                         & \SI{20}{\meter}                                                   & 4.15                                   & 8.81                                   & 3.35                                                    & 4.02                                    & 3.82                                  & 4.53                                & \textbf{3.19}                                       \\
                                                       & \SI{30}{\meter}                                                   & 5.60                                   & 9.29                                   & 3.99                                                    & 4.89                                    & 4.46                                  & 5.18                                & \textbf{3.82}                                       \\ \hline
                                                       & \SI{10}{\meter}                                                   & 3.92                                   & 4.79                                   & 1.36                                                    & 2.19                                    & 1.67                                  & 2.41                                & \textbf{1.21}                                       \\
outdoor night2                                         & \SI{20}{\meter}                                                   & 5.78                                   & 4.52                                   & 2.42                                                    & 3.15                                    & 2.63                                  & 3.24                                & \textbf{2.31}                                       \\
                                                       & \SI{30}{\meter}                                                   & 7.05                                   & 5.34                                   & 3.47                                                    & 3.92                                    & 3.58                                  & 3.47                                & \textbf{3.28}                                       \\ \hline
                                                       & \SI{10}{\meter}                                                   & 4.15                                   & 4.64                                   & 1.20                                                    & 2.86                                    & 1.42                                  & 2.21                                & \textbf{1.01}                                       \\
outdoor night3                                         & \SI{20}{\meter}                                                   & 6.00                                   & 4.08                                   & 2.44                                                    & 4.46                                    & \textbf{2.33}                         & 3.31                                & 2.34                                                \\
                                                       & \SI{30}{\meter}                                                   & 7.24                                   & 4.93                                   & 3.64                                                    & 5.05                                    & \textbf{3.18}                         & 4.72                                & 3.43                                                \\ \hline
\end{tabular}
}\vspace{-1ex}
\end{table*}

\begin{figure*}[t!]
    \centering
    \resizebox{0.8\textwidth}{!}{%
    \begin{tabular}{ccc}
        \includegraphics[height=0.15\linewidth]{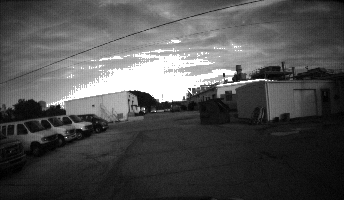}&              \includegraphics[height=0.15\linewidth]{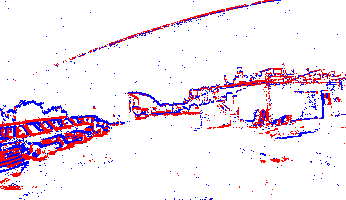}&        \includegraphics[height=0.15\linewidth]{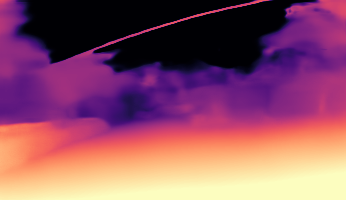}\\
        \footnotesize{(a) Image input} & \footnotesize{(b) Event input} & \footnotesize{(c) E2Depth $[S \to (S+R)]$} \\
        \includegraphics[height=0.15\linewidth]{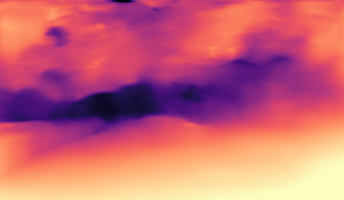} &
        \includegraphics[height=0.15\linewidth]{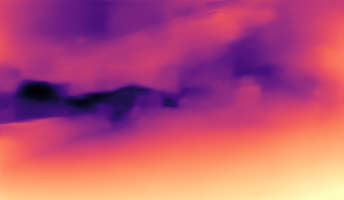} &
        \includegraphics[height=0.15\linewidth]{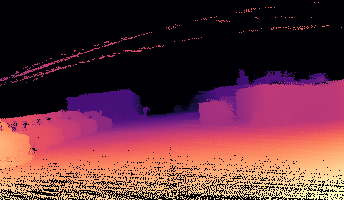} \\
        \footnotesize{(d) I baseline $[S \to R]$} & \footnotesize{(e) \statenet{} $[S \to R]$}  & \footnotesize{(f) Ground truth depth map}
    \end{tabular}}
    \caption{Qualitative performance on the MVSEC dataset. (a) and (b) show the image and event input. (c) Results of E2Depth ~\cite{Hidalgo20threedv} when trained on both simulation and real-world data simultaneously. This training configuration allows to achieve qualitative better results for the sky. However, it was not applicable to our network as EventScape and MVSEC are recorded with different intrinsic camera parameters. (d) \& (e) I baseline and \statenet{} retrained on real-world data. The sky is predicted wrong, however, objects on the ground are correct which is reflected in the superior metrics compared to E2Depth.}
    \label{fig:pictures_qualitative_MVSEC}
    \vspace{-2ex}
\end{figure*}

To test the generalization of our network to real data we first test on four \emph{outdoor\_night} sequences and the \emph{outdoor\_day1} sequence, without retraining and report the mean absolute depth error between the prediction and ground truth depth maps, indicated by [S] in Tab. \ref{tab:results_mvsec}. 
Methods fine-tuned on the MVSEC dataset (as was done in \cite{Hidalgo20threedv}) are indicated with [S$\rightarrow$R].
We compare against different event-based~\cite{Hidalgo20threedv,Zhu19cvpr} and frame-based ~\cite{Li18cvpr} state-of-the-art methods as well as our image-based baseline. 
We evaluate the error for ground truth depth up to \SI{10}{\meter},  \SI{20}{\meter} and  \SI{30}{\meter}, only considering pixels with valid depth measurements. 
An example of the ground truth depth map is shown in Fig. \ref{fig:pictures_qualitative_MVSEC}.
For the event-based baselines, events are grouped into a single packet between depth labels.
While~\cite{Zhu19cvpr} processes them in a feed-forward network,~\cite{Hidalgo20threedv} uses a recurrent architecture. \rev{Similar to our method, \cite{Hidalgo20threedv} pretrains on synthetic events from the DENSE dataset, recorded in CARLA.}
Since both event-based methods are trained on \emph{outdoor\_day2} we did not include it for evaluation.
The frame-based methods on the other hand only process the images and ~\cite{Li18cvpr} uses a feed-forward method.
Since the method ~\cite{Li18cvpr} is trained on images from the internet (the MegaDepth Dataset~\cite{Li18cvpr}) we include the recurrent image-baseline from Sec. \ref{sec:eventscape}. 
For image-based methods, we evaluate only on predictions with a temporal misalignment with depth maps of at most 10 ms.
\newline%
\indent To improve the prediction quality, we fine-tune our network and the I baseline on the \emph{outdoor\_day2} sequence from MVSEC, as was done in ~\cite{Hidalgo20threedv}. 
We choose different values $\alpha=3.7$ and $D_\text{max}$ to normalize depth and unroll the sequence a by 8 steps due to memory limitations. 
Apart from these changes, we train with the same data configurations as during testing.
Again one step is defined as the data measured between two depth labels. 
However, while in EventScape two losses could be used per label due to image and depth map alignment, here we only use a single loss.
\newline
\newline
\noindent\textbf{Results:} For methods only trained on synthetic data, (indicated with $[S]$), \statenet{} generalizes better than I. 
However, sim2real results are significantly worse than E2Depth~\cite{Hidalgo20threedv}. 
The results for $[S \to R]$ are significantly better for both I and \statenet{}. 
The high performance of the I baseline shows the importance of image data for depth prediction by outperforming E2Depth~\cite{Hidalgo20threedv}.
However, as noted, purely frame-based methods cannot generate predictions in the blind-time between frames.
Finally, \statenet{} outperforms all other methods on almost all datasets. 
\omitifnotinarxiv{We compare the qualitative predictions of \statenet{} (Fig. \ref{fig:pictures_qualitative_MVSEC}) and show more comparisons in Fig. 2 and Sec. 1-A of the appendix.}
\newline%
\indent Compared to E2Depth~\cite{Hidalgo20threedv}, our method can reconstruct more details and we observe that there are artifacts in the sky.
This is because during training these regions are masked since they do not have valid LiDAR depth.
In fact, the I baseline also suffers from these artifacts. 
We thus also think that these artifacts are due to the difference of images in the training and test set. 
While in the training set the sky is clear, in the test set the sky is overcast, so the network mistakenly thinks it is structure. 
Purely event-based methods such as E2Depth~\cite{Hidalgo20threedv} do not rely on images, and thus do not suffer from these issues, however, their accuracy in other areas of the image is reduced as indicated by Tab. \ref{tab:results_mvsec}.
Training on more realistic data could significantly improve sky predictions. 
Furthermore, the MVSEC depth labels could be improved by using semantic segmentation to detect the sky and adding maximum depth to the depth labels at these pixels. 
Another approach would be using multitask learning with depth estimation and semantic segmentation to mask the sky at test time.

\subsection{Timing Results}
We measure the processing speed of our architecture on a Quadro RTX 8000 GPU. 
While processing a voxel grid takes 3.08 ms, processing an image takes 3.14 ms, resulting in a throughput of up to 320 Hz.  
\section{Conclusion}
In this work, we introduce \statenet{}, a novel network architecture which generalizes traditional RNNs to 
handle asynchronous and irregular data from multiple sensors.
We applied \statenet{} to the task of monocular depth estimation, where we showed that it can make high-quality predictions in the blind-time between image frames, as shown by experiments on simulated data from EventScape and real data from MVSEC. 
On both synthetic and real datasets, \statenet{} outperforms unimodal baselines by leveraging the complementarity of events and frames.
Furthermore, it also outperforms several baselines and state-of-the-art method based on classical RNNs. 
This is because our network is more robust to changes to the input frequency. 
Finally, to enable further reasearch on multi-modal learning and depth predictions for events we release EventScape, a synthetic large-scale dataset recorded in the CARLA simulator, comprising events, frames, semantic labels, depth maps, and vehicle navigation parameters, to enable further research on multimodal learning using events.
This work makes an important step towards solving the problem of fusing asynchronous and multimodal sensory data for learning.

\begin{table*}[]
\centering
\caption{\label{tab:statenet_details} Architecture details of \statenet{}}
\begin{tabular}{c|cccccc}
\hline
Module                          & Layer type   & Input channels & Output channels & Kernel & Stride & Padding \\ \hline
\multirow{4}{*}{Encoder Images} & Conv         & 1              & 32              & 5      & 1      & -       \\
                                & Conv         & 32             & 64              & 5      & 2      & -       \\
                                & Conv         & 64             & 128             & 5      & 2      & -       \\
                                & Conv         & 128            & 256             & 5      & 2      & -       \\ \hline
\multirow{4}{*}{Encoder Events} & Conv         & 5              & 32              & 5      & 1      & -       \\
                                & Conv         & 32             & 64              & 5      & 2      & -       \\
                                & Conv         & 64             & 128             & 5      & 2      & -       \\
                                & Conv         & 128            & 256             & 5      & 2      & -       \\ \hline
State combination stage 1       & ConvGRU      & 128            & 64              & 3      & 1      & -       \\
State combination stage 2       & ConvGRU      & 256            & 128             & 3      & 1      & -       \\
State combination stage 3       & ConvGRU      & 512            & 256             & 3      & 1      & -       \\ \hline
\multirow{3}{*}{Residual block} & Conv         & 256            & 256             & 3      & 1      & -       \\
                                & Conv         & 256            & 256             & 3      & 1      & -       \\
                                & Conv         & 256            & 256             & 3      & 1      & -       \\ \hline
\multirow{4}{*}{Decoder}        & UpsampleConv & 256            & 128             & 5      & 1      & 2       \\
                                & UpsampleConv & 128            & 64              & 5      & 1      & 2       \\
                                & UpsampleConv & 64             & 32              & 5      & 1      & 2       \\
                                & Conv         & 32             & 1               & 1      & 1      & -       \\ \hline
\end{tabular}

\end{table*}

\begin{table*}[]
\centering
\caption{\label{tab:baseline_details} Architecture details of event-based (E), frame-based (I), and event and frame-based (E+I) baselines.}

    \begin{tabular}{c|cccccc}
    \hline
    Module                          & Layer type   & Input channels    & Output channels & Kernel & Stride & Padding \\ \hline
    \multirow{4}{*}{Encoder}        & Conv         & I: 1, E: 5, E+1:6 & 32              & 5      & 1      & -       \\
                                    & ConvLSTM         & 32                & 64              & 5      & 2      & -       \\
                                    & ConvLSTM         & 64                & 128             & 5      & 2      & -       \\
                                    & ConvLSTM         & 128               & 256             & 5      & 2      & -       \\ \hline
    \multirow{3}{*}{Residual block} & Conv         & 256               & 256             & 3      & 1      & -       \\
                                    & Conv         & 256               & 256             & 3      & 1      & -       \\
                                    & Conv         & 256               & 256             & 3      & 1      & -       \\ \hline
    \multirow{4}{*}{Decoder}        & UpsampleConv & 256               & 128             & 5      & 1      & 2       \\
                                    & UpsampleConv & 128               & 64              & 5      & 1      & 2       \\
                                    & UpsampleConv & 64                & 32              & 5      & 1      & 2       \\
                                    & Conv         & 32                & 1               & 1      & 1      & -       \\ \hline
    \end{tabular}
\end{table*}


\section{Appendix}
In Sec.\ref{app:sec:additional_qualitative} we show additional qualitative results of our method on the MVSEC dataset.
We further analyze the dependence of our method on the input modality in Sec.\ref{app:sec:dep_input_modality} and offer more details on the network architecture of RAM Net and the event-based, frame-based and event and frame-based baselines in Sec.\ref{app:sec:arch}

\subsection{Qualitative Results}
\label{app:sec:additional_qualitative}
In the last row of Fig.\ref{app:fig:qualtitative} we show additional qualitative results of our method on the MVSEC dataset.
We compare against E2Depth\cite{Hidalgo20threedv} (second row) and MegaDepth\cite{Li18cvpr} (fifth row). 
Ground truth depth from the LiDAR is displayed in the third row.

\subsection{Dependence on Input Modalities}
\label{app:sec:dep_input_modality}
\begin{figure*}[t!]
    \centering
    \begin{tabular}{cccc}
        \includegraphics[height=0.15\linewidth]{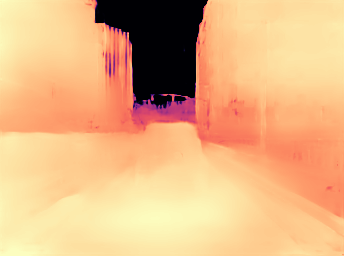} &
        \includegraphics[height=0.15\linewidth]{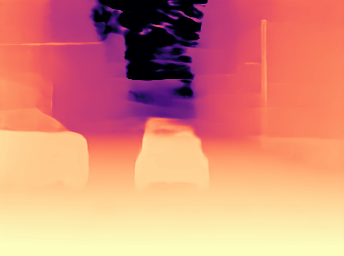} &
        \includegraphics[height=0.15\linewidth]{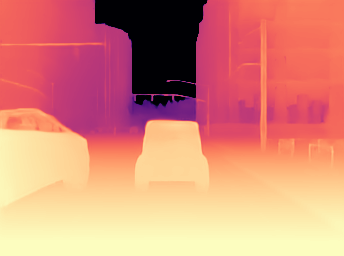}&
        \includegraphics[height=0.15\linewidth]{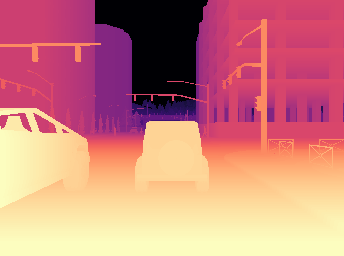}\\
        \footnotesize{(a) \statenet{} images-only} & \footnotesize{(b) \statenet{} events-only}  & \footnotesize{(c) \statenet{} images \& events}&\footnotesize{(d) Ground truth depth map}
    \end{tabular}
    \caption{Ablation studies with different inputs to \statenet{}. No retraining has been conducted for these configurations, all use the trained model at 5Hz image frequency and 25Hz event frequency. At test time, one of these modalities was omitted. (d) For image-only inputs the network sometimes falsely directly translated intensity changes to depth changes, like at the building on the left. (e) For event only, this might occur less as the network has to rely on geometrical cues instead of intensities. However, the events-only prediction suffers from artefacts in the sky and blurred edges of buildings. (f) The prediction with both input modalities shows that \statenet{} is able to include the advantages of both modalities for an ideal depth prediction.}
    \label{fig:pictures_ablationstudies}
\end{figure*}

\begin{figure*}[t!]
    \centering
    \begin{tabular}{cccc}
    \includegraphics[width=0.25\textwidth,keepaspectratio]{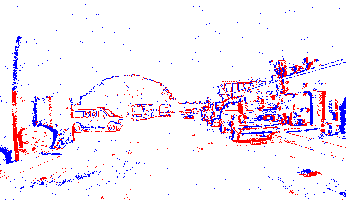}&
    \includegraphics[width=0.25\textwidth,keepaspectratio]{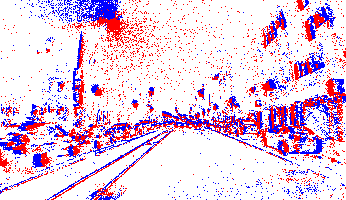}&
    \includegraphics[width=0.25\textwidth,keepaspectratio]{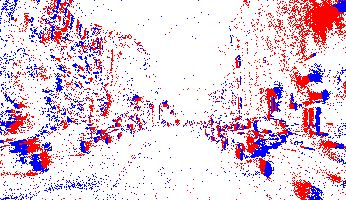}&
    \includegraphics[width=0.25\textwidth,keepaspectratio]{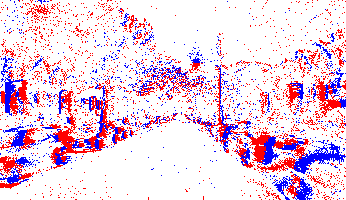}\\
    \includegraphics[width=0.25\textwidth,keepaspectratio]{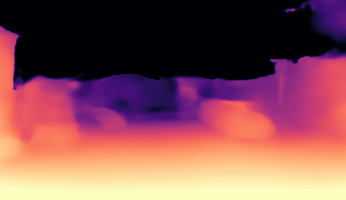}&
    \includegraphics[width=0.25\textwidth,keepaspectratio]{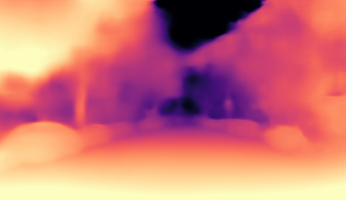}&
    \includegraphics[width=0.25\textwidth,keepaspectratio]{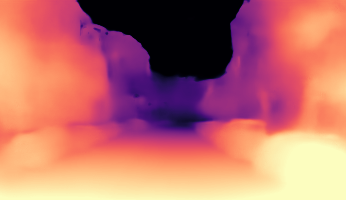}&
    \includegraphics[width=0.25\textwidth,keepaspectratio]{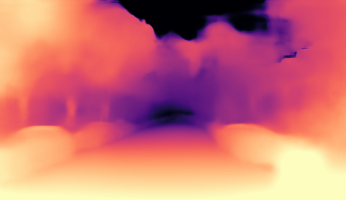}\\
    \includegraphics[width=0.25\textwidth,keepaspectratio]{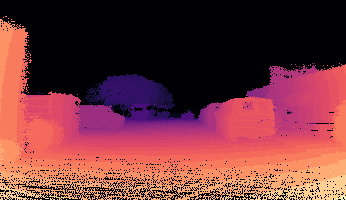}&
    \includegraphics[width=0.25\textwidth,keepaspectratio]{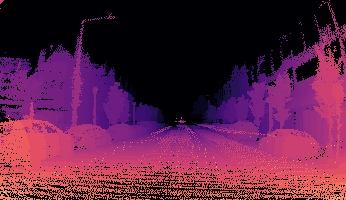}&
    \includegraphics[width=0.25\textwidth,keepaspectratio]{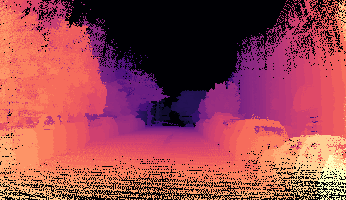}&
    \includegraphics[width=0.25\textwidth,keepaspectratio]{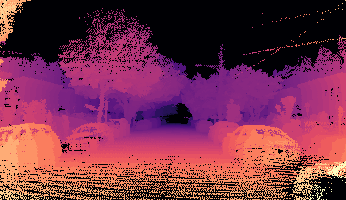}\\
    \includegraphics[width=0.25\textwidth,keepaspectratio]{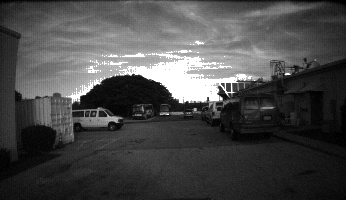}&
    \includegraphics[width=0.25\textwidth,keepaspectratio]{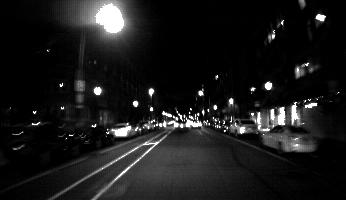}&
    \includegraphics[width=0.25\textwidth,keepaspectratio]{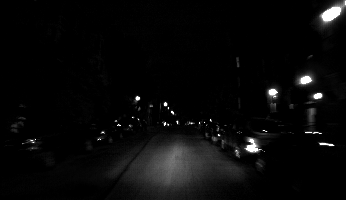}&
    \includegraphics[width=0.25\textwidth,keepaspectratio]{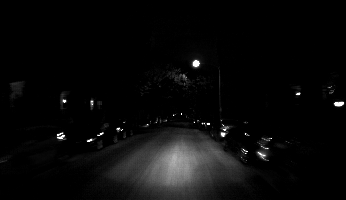}\\
    \includegraphics[width=0.25\textwidth,keepaspectratio]{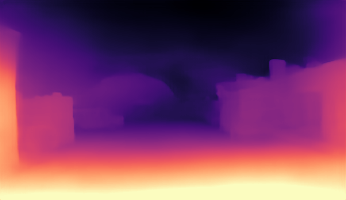}&
    \includegraphics[width=0.25\textwidth,keepaspectratio]{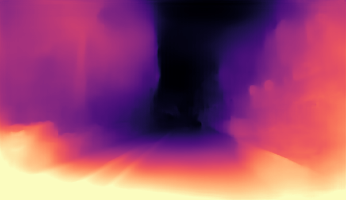}&
    \includegraphics[width=0.25\textwidth,keepaspectratio]{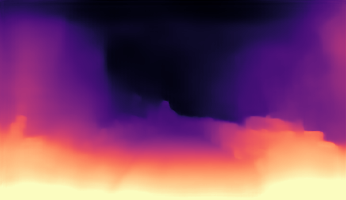}&
    \includegraphics[width=0.25\textwidth,keepaspectratio]{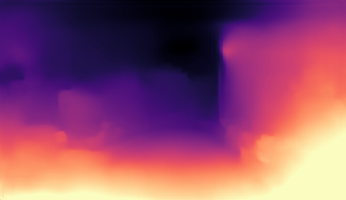}
    \\
    \includegraphics[width=0.25\textwidth,keepaspectratio]{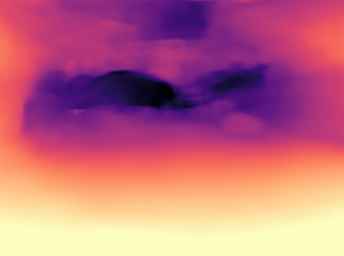}&
    \includegraphics[width=0.25\textwidth,keepaspectratio]{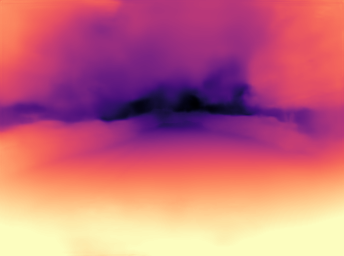}&
    \includegraphics[width=0.25\textwidth,keepaspectratio]{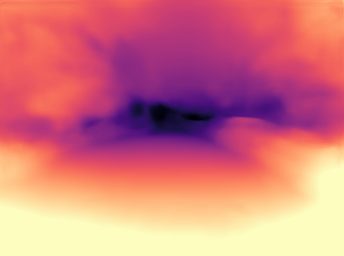}&
    \includegraphics[width=0.25\textwidth,keepaspectratio]{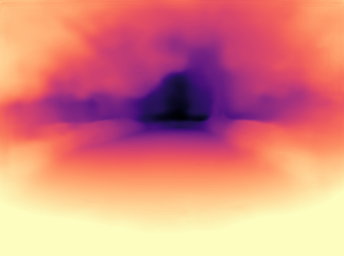}\\
    \emph{outdoor\_day1}&
    \emph{outdoor\_night1}&
    \emph{outdoor\_night2}&
    \emph{outdoor\_night3}
    \end{tabular}
    \caption{Qualitative results of our method on the four test sequences of the MVSEC dataset (last row).
    The first row shows a rendering of the events, the second row shows predictions by E2Depth\cite{Hidalgo20threedv}. 
    The third row shows the ground truth depth from the LiDAR and the fourth row shows the frames. The fifth row show the frame-based method MegaDepth\cite{Li18cvpr} prediction on the frames. From ~\textit{outdoor\_day1},~\textit{outdoor\_night1}, ~\textit{outdoor\_night2} and ~\textit{outdoor\_night3} we use the samples $3741$, $288$, $2837$ and $2901$ respectively.}
   \label{app:fig:qualtitative}
\end{figure*}

To further understand how RAM Net uses the inputs from individual sensors to make predictions we use the network trained on EventScape with both input modalities present and investigate how the predictions change if we omit events or frames.
Fig. \ref{fig:pictures_ablationstudies} shows qualitative depth map predictions of \statenet{} when we omit events (a) or frames (b). 
We observe that when presented with events only (b) the network fails to correctly predict the sky at some positions and the overall sharpness of the depth predictions is reduced. 
That indicates that the image data helps to distinguish object borders from the sky. 
On the other hand, when presented with images only (a), the network mistakenly includes intensity differences in the depth prediction, as seen in the building on the left. 
We thus hypothesize that events provide useful geometric structure queues to the network which are missing in (a).
By omitting one of the input modalities completely, interesting conclusions can be drawn on how the network works. 
The experiment shows that \statenet{} can incorporate the strong points of both modalities for an ideal depth prediction.
\subsection{Architecture details}
\label{app:sec:arch}
The network architecture of RAM Net and the event-based, frame-based, and event and frame-based baselines can be seen in Tab.\ref{tab:statenet_details} and Tab.\ref{tab:baseline_details} respectively. 
The baseline methods differ only in the number of input channels of the input convolution. 
They have the same decoder, residual block structure as RAM Net.
While RAM Net has a non-recurrent encoder with recurrent state combination blocks, the baselines feature ConvLSTMs at the encoder stage.

\setlength{\tabcolsep}{0pt}

\bibliographystyle{IEEEtran}
\bibliography{all.bib}
\end{document}